\documentclass[11pt,a4paper]{article}
\usepackage[hyperref]{acl2020}
\usepackage{times}
\usepackage{latexsym}
\usepackage{appendix}

\usepackage{microtype}

\aclfinalcopy % Uncomment this line for the final submission
\def\aclpaperid{872} %  Enter the acl Paper ID here

\usepackage{url}
\usepackage{amssymb}
\usepackage{bbm}
\usepackage{multirow}
\usepackage{stfloats}
\usepackage{graphicx}
\usepackage{booktabs}
\usepackage{xspace}

\usepackage{algorithm}  
\usepackage{algpseudocode}  
\usepackage{amsmath}  
\usepackage[english]{babel}

\DeclareMathOperator*{\argmin}{argmin}
\DeclareMathOperator*{\argmax}{argmax}

\title{A Self-Training Method for Machine Reading Comprehension with \\Soft Evidence Extraction}
\author{Yilin Niu$^1$\footnotemark[1] , Fangkai Jiao$^2$\footnotemark[1] , Mantong Zhou$^1$ , Ting Yao$^3$ , Jingfang Xu$^3$ , Minlie Huang$^1$\footnotemark[2]\\
  $^1$ Department of Computer Science and Technology, Institute for Artificial Intelligence, \\
  State Key Lab of Intelligent Technology and Systems,\\
  Beijing National Research Center for Information Science and Technology,\\
  Tsinghua University, Beijing 100084, China\\ 
  $^2$ School of Computer Science and Technology, Shandong University \\
  $^3$ Sogou Inc., Beijing, China \\
  {\small \tt niuyl14@tsinghua.org.cn}
  \quad {\small \tt jiaofangkai@hotmail.com}
  \quad {\small \tt zmt.keke@gmail.com}\\
  {\small \tt \{yaoting,jingfang\}@sogou-inc.com} \quad 
  {\small \tt aihuang@tsinghua.edu.cn}}

\date{}

\newcommand{\methodname}{STM\xspace}

\newcommand{\BERTMLP}{BERT-MLP\xspace}
\newcommand{\BERTHAtt}{BERT-HA\xspace}
\newcommand{\BERTHAttRule}{BERT-HA+Rule\xspace}
\newcommand{\BERTHAttRL}{BERT-HA+RL\xspace}
\newcommand{\BERTHAttGold}{BERT-HA+Gold\xspace}
\newcommand{\BERTSelfCTM}{BERT-HA+STM\xspace}

\begin{document}
\maketitle
\renewcommand{\thefootnote}{\fnsymbol{footnote}}
\footnotetext[1]{Equal contribution}
\footnotetext[2]{Corresponding author: Minlie Huang.}
\renewcommand{\thefootnote}{\arabic{footnote}}
\begin{abstract}

% 任务定义
%%% aims to answer questions according to some reference text.
% 现有工作的一般做法
Neural models have achieved great success on machine reading comprehension (MRC), many of which typically consist of two components: an evidence extractor and an answer predictor. The former seeks the most relevant information from a reference text, while the latter is to locate or generate answers from the extracted evidence. Despite the importance of evidence labels for training the evidence extractor, they are not cheaply accessible, particularly in many non-extractive MRC tasks such as YES/NO question answering and multi-choice MRC.

To address this problem, we present a Self-Training method (STM), which supervises the evidence extractor with auto-generated evidence labels in an iterative process. At each iteration, a base MRC model is trained with golden answers and noisy evidence labels. The trained model will predict pseudo evidence labels as extra supervision in the next iteration. We evaluate STM on seven datasets over three MRC tasks. Experimental results demonstrate the improvement on existing MRC models, and we also analyze how and why such a self-training method works in MRC. The source code can be obtained from \href{https://github.com/SparkJiao/Self-Training-MRC}{https://github.com/SparkJiao/Self-Training-MRC}
\end{abstract}

\section{Introduction}

% 简要介绍我们研究的任务，non-extractive MRC
%%%Recently, there is more and more attention on Machine Reading Comprehension (MRC). 
Machine reading comprehension (MRC) has received increasing attention recently, which can be roughly divided into two categories: extractive and non-extractive MRC. Extractive MRC requires a model to extract an answer span to a question from reference documents, such as the tasks in SQuAD~\citep{squad} and CoQA~\citep{CoQA}. In contrast, non-extractive MRC infers answers based on some \textbf{evidence} in reference documents, including 
%abstractive MRC~\citep{MARCO}, %%%%generative??? % 查了一些说法，感觉abstractive更常见一点
Yes/No question answering~\citep{BoolQ},
multiple-choice MRC~\citep{race,multirc,dream}, and open domain question answering~\citep{quasar}. 
As shown in Table \ref{tab:examples}, evidence plays a vital role in MRC~\citep{GEAR,CognitiveGraph,hardselector}, and the
%%%%%%% widely: 这种方法现在确实有不少人在提，这里补充一些引用 --nyl
coarse-to-fine paradigm has been widely adopted in multiple models~\citep{coarse-to-fine,HumanLikeReadingStrategy,R3} where an evidence extractor first seeks the evidence from given documents and then an answer predictor infers the answer based on the evidence.
%Evidence plays a vital role in MRC~\citep{GEAR,CognitiveGraph,hardselector}, 
However, it is challenging to learn a good evidence extractor since there lack evidence labels for supervision. 

Manually annotating the golden evidence is expensive.
Therefore, some recent efforts have been dedicated to improving MRC by leveraging noisy evidence labels when training the evidence extractor. Some works~\citep{DS-QA,hardselector} generate distant labels using hand-crafted rules and external resources. Some studies~\citep{R3,coarse-to-fine} adopt reinforcement learning (RL) to decide the labels of evidence, however such RL methods suffer from unstable training. More distant supervision techniques are also used to refine noisy labels, such as deep probability logic~\citep{MMRCselector}, but they are hard to transfer to other tasks. Nevertheless, improving the evidence extractor remains challenging when golden evidence labels are not available.

\begin{table}[ht]
\begin{tabular}{lp{0.4\textwidth}}
\toprule[1pt]
\textbf{Q:} & Did a little boy write the note? \\
\textbf{D:} & ...\textbf{This note is from a little girl}. She wants to be                   your friend. If you want to  be her friend, ...\\
\textbf{A:} & No  \\
\hline
\textbf{Q:} & Is she carrying something?\\
\textbf{D:} & ...On the step, I find the elderly Chinese lady, small and slight, holding the hand of a little boy. \textbf{In her other hand, she} \textbf{holds a paper carrier bag.} ...\\
\textbf{A:} & Yes \\ 
\bottomrule[1pt]
\end{tabular}
\caption{Examples of Yes/No question answering. Evidential sentences in bold in reference documents are crucial to answer the questions.}
\label{tab:examples}
\end{table} %%%为甚给三个例子，是不是两个就好？这几个例子有什么不同特点吗？？？

In this paper, we present a general and effective method based on Self-Training~\citep{selftraining}  to improve MRC with soft evidence extraction when golden evidence labels are not available. 
%%%Co-Training is a widely used method in semi-supervised learning. %%%看看我写的句子是如何衔接的 
%More specifically, our method is based on Self-Training~\citep{selftraining} and works along with base models\footnote{Our method is not constrained on a specific type of base model since it is designed to be a general training method.}, such as BERT~\citep{BERT} with hierarchical attention. 
Following the Self-Training paradigm, a base MRC model is iteratively trained. At each iteration, the base model is trained with golden answers, as well as noisy evidence labels obtained at the preceding iteration. Then, the trained model generates noisy evidence labels, which will be used to supervise evidence extraction at the next iteration. The overview of our method is shown in Figure \ref{fig:co-training}. 
Through this iterative process, evidence is labelled automatically to guide the RC model to find answers, and then a better RC model benefits the evidence labelling process in return.
Our method works without any manual efforts or external information, and therefore can be applied to any MRC tasks.
Besides, the Self-Training algorithm converges more stably than RL.
% For , the final answer is predicted on top of the two trained models.
%%%%%%
% 我们模型的贡献
Two main contributions in this paper are summarized as follows:
%%%这几点要重新梳理一下
\begin{enumerate}
    %%%% 仔细改改，现在逻辑很不通顺
    \item We propose a self-training method to improve machine reading comprehension by soft evidence labeling. 
    %It alleviates the problem of absent evidence annotation in non-extractive MRC tasks. 
    Compared with other existing methods, 
    %such as rule-based and RL-based methods, 
    our method is more effective and general.% without task-specific design and massive external resources.
    
    %\item To the best of our knowledge, this is the first study on exploring self-training for pseudo reasoning processes rather than final labels. %%%%%%%%%% for pseudo reasoning processes的说法会不会有点奇怪，给人的感觉是只是为了生成中间过程而不是提升MRC的最终性能 ---nyl
    % Our method 
    
    \item We verify the generalization and effectiveness of \methodname on several MRC tasks, including Yes/No question answering (YNQA), multiple-choice machine reading comprehension (MMRC), and open-domain question answering (ODQA). Our method is applicable to different base models, including BERT and DSQA~\citep{DS-QA}. Experimental results demonstrate that our proposed method improves base models in three MRC tasks remarkably.
    %%%%%%%%%%%% 这里要不要加上后续分析实验的简要描述，例如对于performance of extractor的实验分析 ---nyl
\end{enumerate}

\section{Related Work}

% MRC的相关工作分类。其中要包括evidence extractor的工作
%%%Recent researches about MRC have several concerns. %%%中国式英语!!!
Early MRC studies focus on modeling semantic matching between a question and a reference document~\citep{BiDAF,FusionNet,SDNet, KnowledgeableReader}.
%%%%On this basis,  中国式英语!!!
%%%further researches enhance matching based models with external knowledge~\citep{KnowledgeableReader}.
%%%知识跟你的论文无关，没有必要强调
In order to mimic the reading mode of human, hierarchical coarse-to-fine methods are proposed~\citep{coarse-to-fine,HumanLikeReadingStrategy}. Such models first read the full text to select relevant text spans, and then infer answers from these relevant spans.
Extracting such spans in MRC is drawing more and more attention, though still quite challenging~\citep{MMRCselector}. 
%%%However, extracting such spans in MRC is still difficult and drawing more and more attention~\citep{MMRCselector}.
%%%你这个写法跟前一句对的上吗

% evidence information在哪些任务种很重要
% 在这些任务种evidence extractor是怎么工作的，为什么重要
%%%%For most natural language understanding tasks, it is commonly observed that original input contains much irrelevant information which can distract models. Inspired by such observations, 
Evidence extraction aims at finding evidential and relevant information for downstream processes in a task, which arguably improves the overall performance of the task. Not surprisingly, 
evidence extraction is useful and becomes an important component in fact verification~\citep{GEAR,twowing,UKP,sentEviEmb}, multiple-choice reading comprehension~\citep{MMRCselector,bax2013cognitive,DeducingEvidenceChain}, open-domain question answering~\citep{DS-QA,R3}, multi-hop reading comprehension~\citep{multihopEvidenceExtract,CognitiveGraph}, natural language inference~\citep{BiMPM,ESIM}, and a wide range of other tasks~\citep{whoiskilled,FastAbstractive}.

% 之前的evidence extractor都是怎么设计的、怎么训练的
% 两种设计方式：soft、hard
% 两种evidence数量：single、multi
% 多种训练方法：Supervised、Unsupervised、RL、weakly-supervised
In general, evidence extraction in MRC can be classified into four types according to the training method. % training methods/training modes？
First, unsupervised methods provide no guidance for evidence extraction~\citep{BiDAF,FlowQA}.
%%%%%%%%% no evidence这个说法合适吗？
Second, supervised methods train evidence extraction with golden evidence labels, which sometimes can be generated automatically in extractive MRC settings~\citep{DS-QA,twowing,UKP}. 
Third, weakly supervised methods rely on noisy evidence labels, where the labels can be obtained by heuristic rules~\citep{hardselector}. Moreover, some data programming techniques, such as deep probability logic, were proposed to refine noisy labels~\citep{MMRCselector}.
Last, if a weak extractor is obtained via unsupervised or weakly supervised pre-training, reinforcement learning can be utilized to learn a better policy of evidence extraction~\citep{R3,coarse-to-fine}.

% 我们的关注点：无evidence label的情况下，如何更好地训练evidence extractor？
% 过去都有哪些类似的工作，他们的问题是什么
% 我们的不同点在哪里，优势在哪里，是否解决了他们的问题
% Co-training简短的工作总结：有很多优秀的有理论依据的co-training变体，我们这里面只使用了最简单的co-training方法，已经取得了不错的结果，使用更优秀的方法会有进一步的提升
For non-extractive MRC tasks, such as YNQA and MMRC, it is cumbersome and inefficient to annotate evidence labels~\citep{sentEviEmb}.
Although various methods for evidence extraction have been proposed, training an effective extractor is still a challenging problem when golden evidence labels are unavailable.
%%%%下面的为什么没有reference？---hml %%%% 这里也可以加，但是引用会和前面重复，这样可以吗---nyl
%%%%%%%%%再尝试找一些引用吧---nyl
\textbf{Weakly supervised methods} either suffer from low performance or rely on too many external resources, which makes them difficult to transfer to other tasks. 
\textbf{RL methods} can indeed train a better extractor without evidence labels. However, they are much more complicated and unstable to train, and highly dependent on model pretraining.
%%%%%去掉吧。。。
%%%%%%%%%%In fact, these two types of methods are able to collaborate to improve evidence extraction in two phases. In the first phase, weakly supervised methods pre-train a weak extractor. In the second phase, RL methods fine-tune the obtained extractor according to some well-designed reward. In this paper, we focus on designing an \textbf{effective} and \textbf{general} weakly-supervised method for the first phase.

%%%%%%%%% 这一段还需要吗
Our method is based on Self-Training, a widely used semi-supervised method.
%%%Proof of traditional Co-Training theory analyzes the requirements which are needed for it to be effective. %%%%
Most related studies follow the framework of traditional Self-Training~\citep{selftraining} and Co-Training~\citep{cotraining}, and focus on designing better policies for selecting confident samples. 
CoTrade~\citep{cotrade} evaluates the confidence of whether a sample has
been correctly labeled via a statistic-based data editing technique~\citep{cut_edge_weight_statistic}.
% CoTrade~\citep{cotrade} estimates confidence via a data editing technology based on cut edge weight statistic~\citep{cut_edge_weight_statistic}. %%%%
%%%% cut edge weight stat???
Self-paced Co-Training~\citep{selfpaced} adjusts labeled data dynamically according to the consistency between the two models trained on different views. 
A reinforcement learning based method~\citep{rlcotraining} designs an additional Q-agent as a sample selector.
%%%%%%%%%Traditional Co-Training is applied in this paper, due to its simplicity and generalization. %%%%%Note that there is room for further improvement if more effective Co-Training methods are used.

\section{Methods}

\subsection{Task Definition and Model Overview}
% 任务形式化定义，引入符号表示
% 在第一次出现方法名称简写的时候加以解释，最后再加这一部分，因为尚不确定在哪里解释比较好
The task of machine reading comprehension can be formalized as follows: given a reference document composed of a number of sentences $D=\{S_1,S_2,\cdots,S_m\}$ and a question $Q$, the model should extract or generate an answer $\hat{A}$ to this question conditioned on the document, formally as
\begin{align} \nonumber
    \hat{A}=\argmax_{A'} P(A'|Q,D).
\end{align}
% 这里需不需要说明Open-domain QA中Si代表的是document
The process can be decomposed into two components, i.e., an evidence extractor and an answer predictor.
%\begin{align}
%    \hat{A}=\argmax_{A'} P_E(E|Q,D)P_A(A'|Q,D,E).\label{eq:infer-answer}
%\end{align}
%%%%%%%%% 这里需要再斟酌一下，因为我们用的是soft evidence extraction，evidence label才是hard的 --nyl
%$E=\{E_1,E_2,\cdots,E_m\}$ denotes the binary evidence labels for each sentence where $0/1$ corresponds to the non-evidence/evidence sentence, respectively.
The golden answer $A$ is given for training the entire model, including the evidence extractor and the answer predictor. %%%%tor是可数，需要定冠词或不定冠词
Denote $E_i$ as a binary evidence label $\{0, 1\}$ for the $i$-th sentence $S_i$, where $0/1$ corresponds to the non-evidence/evidence sentence, respectively.
An auxiliary loss on the evidence labels can help the training of the evidence extractor. %%%%%% 额外的supervision指evidence loss。 nyl

\begin{figure}[ht]
    \centering
    %%%%%%%%%% transfer箭头和两个实线箭头是表示同一个过程，都画出来会不会造成理解的困扰 ---nyl
    \includegraphics[width=0.5\textwidth]{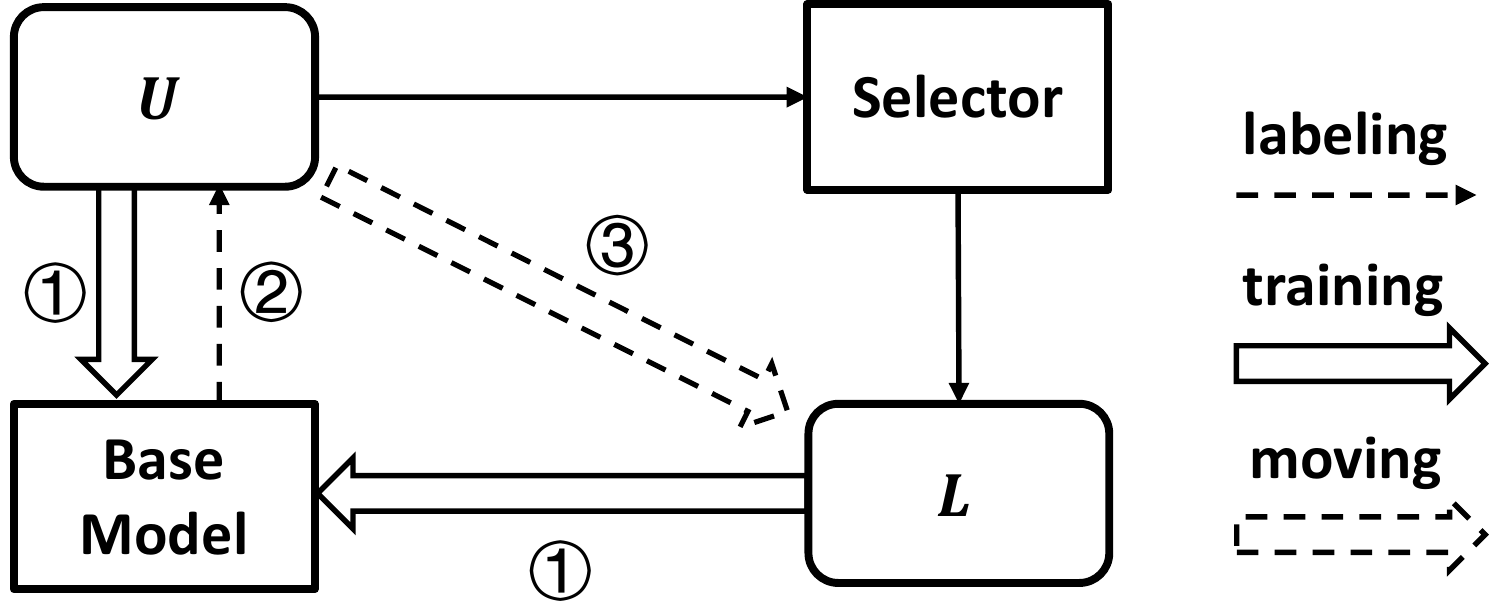}
    \caption{Overview of Self-Training MRC (\methodname). The base model is trained on both $L$ and $U$. 
    After training, the base model is used to generate evidence labels for the data from $U$, and then $Selector$ chooses the most confident samples, which
    will be used to supervise the evidence extractor at the next iteration. The selected data is moved from $U$ to $L$ at each iteration.
    %%%%圆圈圈的编号怎么这么别扭，对应到文字的顺序吧!!!
    %%%The two training arrows from $U_i$ to $Model_i$ are left out in the figure for better presentation.
    }
    %%%%% 实际上在每次训练的时候，只使用L_i中的evidence labels对evidence extractor进行监督，同时使用L_i U_i中的golden answers对整个模型进行监督. nyl
    \label{fig:co-training}
    %%%%%\vspace{-4mm}
\end{figure}

%%%%有几个问题：第一，模型如何启动，是否需要pretraining
%%%%第二，初始标注如何获得，这里提到了golden是指，有些任务中有金标准吗？ % 这里提到的golden answer是指最终的答案
The overview of our method is shown in Figure \ref{fig:co-training}, which is an iterative process. During training, two data pools are maintained and denoted as $U$ (unlabeled data) and $L$ (labeled data). In addition to golden answers, examples in $L$ are annotated with pseudo evidence labels. In contrast, there are only golden answers provided in $U$.
% For $Model_i$ ($i \in \{1,2\}$), denote $U_i$ and $L_i$ as the training sets of instances without and with evidence labels, respectively.
At each iteration, the base model is trained on both data pools (two training arrows). After training, the model makes evidence predictions on unlabeled instances (the labeling arrow), and then $Selector$ chooses the most confident instances from $U$ to provide noisy evidence labels. In particular, the instances with newly generated evidence labels are moved from $U$ to $L$ (the moving arrow), which are used to supervise evidence extraction in the next iteration. 
This process will iterate several times.
% After several rounds, ...

%%%%%%In the following sections, we firstly introduce a typical base model, a BERT-based model (Section \ref{sec:bert-model}). If evidence labels are available, supervision can be imposed on evidence extraction (Section \ref{sec:supervised-extractor}). Our proposed method for improving MRC with soft evidence extraction will be described in Section \ref{sec:our-method}, including generating evidence labels via Co-Training.

\subsection{Base Model}
\label{sec:bert-model}
\begin{figure}[ht]
    \centering
    \includegraphics[width=0.4\textwidth]{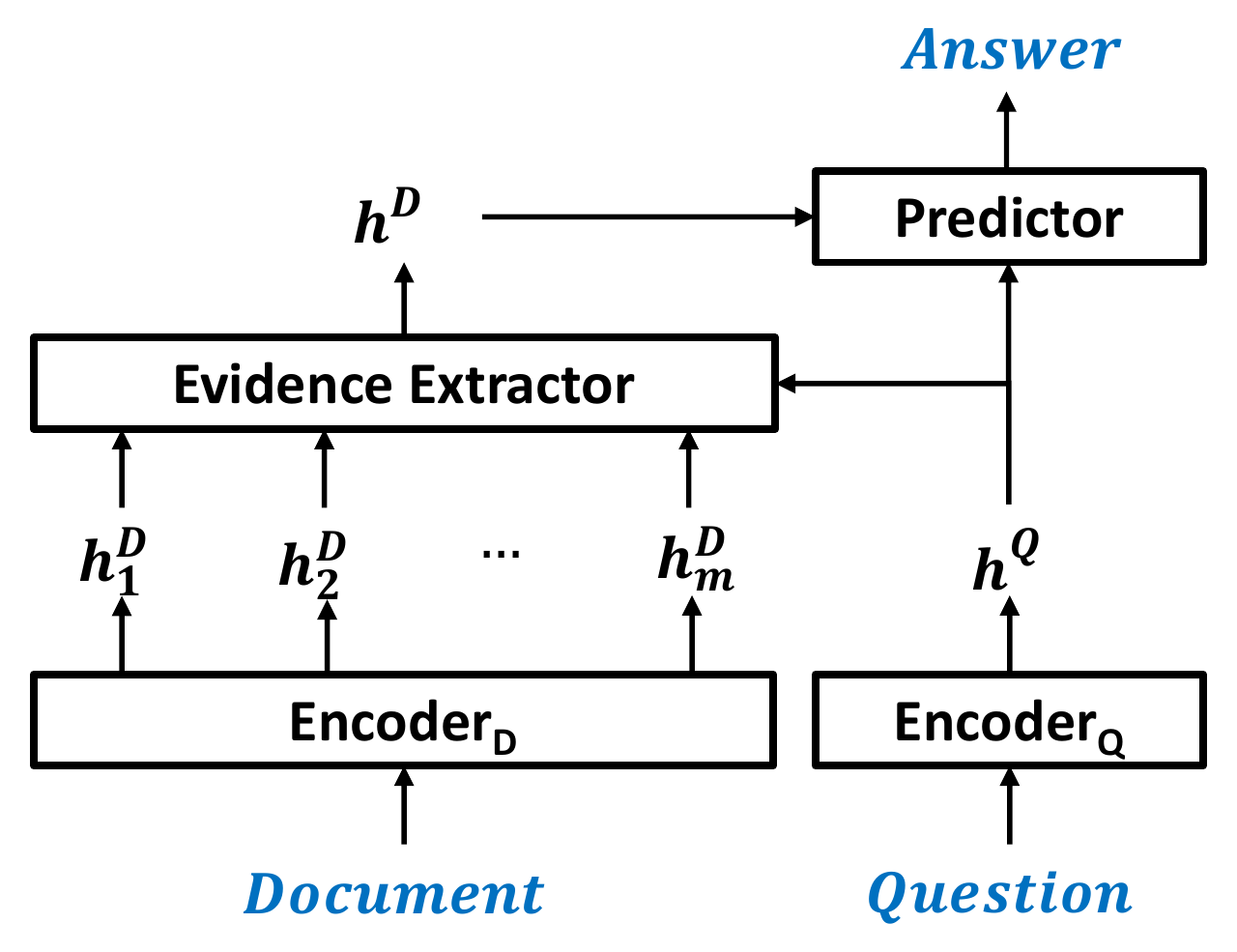}
    \caption{Overall structure of a base model that consists of an encoder layer, an evidence extractor, and an answer predictor. The encoders will obtain $\boldsymbol{h}^Q$ for the question, and $\boldsymbol{h}_i^D$ for each sentence in a document. The summary vector $\boldsymbol{h}^D$ will be used to predict the answer.}
    \label{fig:claim-model}
    %%%%%\vspace{-2mm}
\end{figure}

As shown in Figure \ref{fig:claim-model}, the overall structure of a base model consists of an encoder layer, an evidence extractor, and an answer predictor. 
%To demonstrate the generalization of our self-training method, we employ two types of base models, Transformer-based model~\citep{BERT} and LSTM-based model~\citep{DS-QA}. In the remainder of this section, we illustrate the detailed structure of each module in the Transformer-based model. Due to the length limit, we omit the detailed description of the LSTM-based model.

% 我觉得只要在实验部分简要介绍DSQA就可以了？在方法部分只以bert model为例进行说明。
The encoder layer takes document $D$ and question $Q$ as input to obtain contextual representation for each word. Denote $\boldsymbol{h_{i,j}^D}$ as the representation of the $j$-th word in $S_i$, and $\boldsymbol{h_i^Q}$ as the representation of the $i$-th word in question $Q$.
Our framework is agnostic to the architecture of the encoder, and we show improvements on two widely used encoding models, i.e. Transformer (with BERT, \citealp{BERT}) and LSTM (with DSQA, \citealp{DS-QA}) in the experiments.

The evidence extractor employs hierarchical attention, including token- and sentence-level attention, to obtain the document representation $\boldsymbol{h^D}$. 
%Token-level attention obtains a sentence vector by self-attention~\citep{selfattention} within the words in a sentence.
%Sentence-level attention identifies important sentences conditioned on the question in a \textbf{soft} way.
% 公式格式好丑，改一下

\iffalse 
\noindent\textbf{Bilinear attention.} %%%We use 
%%%a kind of---这不是典型的中国式英语？我们使用了一种xxx 
%%bilinear attention to obtain the matching-based representation. 
The attention score between two vectors is calculated by bilinear attention, as follows:
\begin{align} \nonumber
    &F(\boldsymbol{x},\boldsymbol{y})=f(\boldsymbol{Wx})^\top \boldsymbol{V}\,f(\boldsymbol{Wy}),\label{eq:attention}
\end{align}
where $\boldsymbol{W}$ and $\boldsymbol{V}$ are both parameters, and $f$ is a $ReLU$ function in this paper.
\fi 

\noindent\textbf{Token-level attention} obtains a sentence vector by self-attention~\citep{selfattention} within the words in a sentence, as follows:
\begin{align} \nonumber
    &\boldsymbol{h^D_i}=\sum^{|S_i|}_{j}\alpha_{i,j}\boldsymbol{h^D_{i,j}}, \;\alpha_{i,j}\propto\exp(F^S(\boldsymbol{h^Q},\boldsymbol{h^D_{i,j}})),\\
    &\boldsymbol{s_i^D}=\sum^{|S_i|}_j\beta_{i,j}\boldsymbol{h^D_{i,j}},\;\beta_{i,j}\propto\exp(\boldsymbol{w_sh^D_{i,j}}+b_s), \nonumber
\end{align}
where $\boldsymbol{h^Q}$ is the sentence representation of the question. 
%generated via self-attention~\citep{selfattention} over $\{\boldsymbol{h_1^Q}, \boldsymbol{h_2^Q}, \cdots, \boldsymbol{h_{|Q|}^Q}\}$. 
%$|X|$ denotes the length of a sequence. 
$\alpha_{i,j}$ refers to the importance of word $j$ in sentence $i$, and so on for $\beta_{i,j}$. $\boldsymbol{w_s}$ and $\boldsymbol{b_s}$ are learnable parameters. The attention function $F^S$ follows the bilinear form~\citep{bilinear}.%~\citep{}. bilinear reference? --zmt

\noindent\textbf{Sentence-level attention} identifies important sentences conditioned on the question in a \textbf{soft} way to get the summary vector ($\boldsymbol{h^D}$), as follows:
\begin{align} \nonumber
    &\boldsymbol{h^D}=\sum^m_i\gamma_i \boldsymbol{h^D_i},\;\gamma_i\propto\exp(F^D(\boldsymbol{h^Q},\boldsymbol{s_i^D})),
\end{align}
%%%%
%%%%%把下面大段的符号解释分散到每个公式里面去！！！readabilities多差呢，这么搞的话 
where
%$F^S(\boldsymbol{x},\boldsymbol{y})$ and $F^D(\boldsymbol{x},\boldsymbol{y})$ have the same form  of Equation \ref{eq:attention}, but are parameterized differently;
$F^D$ has the same bilinear form as $F^S$ with different parameters.
%$m$ is the number of sentences contained in $D$; and
$\gamma_i$ refers to the importance of the corresponding sentence.

% 这一部分看看怎么说比较好，可能有点难以统一各个MRC任务
% different MRC tasks ***have*** 这句话有没有好的表达方式：给每个MRC task配备了不同的predictor？ 
The answer predictor adopts different structures for different MRC tasks. For Yes/No question answering, we use a simple linear classifier to infer answers. For multiple choice MRC, we use a Multiple Layer Perceptron (MLP) with Softmax to obtain the score of each choice.
%%%%get是口语化的词
% For Yes/No question answering and multiple-choice MRC, we use Multiple Layer Perceptron (MLP) as a classifier.
And for open-domain question answering, one MLP is used to predict the answer start, and another MLP is used to predict the end.%%%joint---谁和谁的joint？？？---hml %%%%%这里想表达的是两个MLP组成了这个predictor，也就是两个MLP的组合 ---nyl
%%%%%
% We choose ReLU as activation function in MLP.

\subsection{Loss Function}\label{sec:supervised-extractor}
We adopt two loss functions, one for task-specific loss and the other for evidence loss.

The task-specific loss is defined as the negative log-likelihood (NLL) of predicting golden answers, formally as follows:
% 公式有问题，改一下
\begin{align}  \nonumber
    &L_A(D,Q,A)=-\log P(\hat{A}=A|D,Q),
\end{align}
%%%%%%%%%不要换行！！！---hml 
%%%%where $\hat{A}$ is from Equation \ref{eq:infer-answer}. %%% 为什么要别人去看前面，累不累？？？
where $\hat{A}$ denotes the predicted answer and $A$ is the golden answer.

% 这一部分可能可以作为实验部分的实现细节
% $P_i$ refers to the probability of the $i$-th class. We splice document and question together as input of BERT, which is a common operation in other BERT based models for MRC.
% % 需要仔细考虑一下common operation这个说法是否准确
% Therefor, $h_D$ contains information of both $D$ and $Q$, so that $h^Q$ is not a necessary input for answer predictor.

When the evidence label $E$ is provided, we can impose supervision on the evidence extractor. For the most general case, we assume that a variable number of evidence sentences exist in each sample $(Q,A,D)$. 
Inspired by previous work~\citep{multihopEvidenceExtract} that used multiple evidences, we calculate the evidence loss step by step. Suppose we will extract $K$ evidence sentences. At the first step, we compute the loss of selecting the most plausible evidence sentence. At the second step, we compute the loss in the remaining sentences, where the previously selected sentence is \emph{masked} and not counted in computing the loss at the second step. The overall loss is the average of all the step-by-step loss until we select out $K$ evidence sentences. In this manner, we devise a BP-able surrogate loss function for choosing the top $K$ evidence sentences.

Formally, we have 
%%%%%We define the timestep to be the operation to extract an evidence sentence, which lead to the minimum loss at the current timestep. 
%%%The final evidence loss is defined as the average evidence loss across all timesteps:
\begin{align}  \nonumber
    &L_E(D, Q, E)=\frac{1}{K}\sum_{k=1}^K H(D,Q,E,M^{k}),
\end{align}
where $K$ is the number of evidence sentences, a pre-specified hyperparamter. $M^k=\{M_1^k,M_2^k,\cdots,M_m^k\}$ and each $M_i^k \in \{0, -\infty \}$ is a sentence mask, where $0$ means sentence $i$ is \emph{not} selected before step $k$, and $-\infty$ means selected. 

%%%$H(D,Q,E,M^{k})$ calculates the evidence loss at timestep $t$. 
At each step, the model will compute an attention distribution over the unselected sentences, as follows:
\begin{align} \nonumber
    &\lambda_i^k = \frac{\exp(F^D(\boldsymbol{h^Q},\boldsymbol{s_i})+M_i^{k})}
    {\sum_j(\exp(F^D(\boldsymbol{h^Q},\boldsymbol{s_j})+M_j^{k}))}.
\end{align}
As $M_i^{k}=-\infty$ for the previously selected sentences, the attention weight on those sentences will be zero, in other words, they are masked out. Then, the step-wise loss can be computed as follows: 
\begin{align} \nonumber
    &H(D,Q,E,M^{k})=- \log \max_i (\lambda_i^{k}*E_i),
\end{align}
where $\lambda_i^k$ indicates the attention weight for sentence $i$, and $E_i\in \{0,1\}$ is the evidence label for sentence $i$.
%%%这个E_i怎么得来的！！！
The sentence with the largest attention weight will be chosen as the $k$-th evidence sentence. 

%%%%%%%%%%%  E_i 是什么？？？ 把这两个公式ide物理意义说清楚，不行
%%%%%%%%%%%where $\lambda_i^t$ represents the importance of $S_i$ at timestep $t$. $E_i$ is the evidence label with binary value $0/1$ for the $i$-th sentence. $\lambda_i^t$ is calculated in a similar way to $\gamma_i$ in Equation \ref{eq:sentence-attention} without regard to $M_i^{t}$. $M_i^{t}$ is a binary mask with values of $0$ or $-\infty$. $M_i^{t}=-\infty$ indicates that $S_i$ is not considered when calculating the evidence loss at timestep $t$. 

For each sentence $i$, $M_i^1$ is initialized to be $0$. At each step $k(k>1)$, the mask $M_i^k$ will be set to $-\infty$ if sentence $i$ is chosen as an evidence sentence at the preceding step $k-1$, and the mask remains unchanged otherwise. Formally, the mask is updated as follows:
\begin{align} \nonumber
    &M_i^{k}=\left\{
        \begin{array}{ll}
        -\infty       &      {i=\argmax\limits_j (\lambda_j^{k-1}E_j)}\\
        M_i^{k-1}          & otherwise
        \end{array} \right..
\end{align}

During training, the total loss $L$ is the combination of the task-specific loss and the evidence loss:
\begin{align}\label{eq:total-loss}
    % \vspace{-1mm}
            L=\sum_{(D,Q,A)\in U \cup L}&L_A(D,Q,A)+\notag\\
              \eta \sum_{(D,Q,E)\in L}&L_E(D,Q,E),
\end{align}
where $\eta$ is a factor to balance the two loss terms. $L$ and $U$ denote the two sets in which instances with and without evidence labels, respectively. Note that the evidence label in $L$ is automatically obtained in our self-training method. 
% Evidence extractor supervised by golden evidence labels provides an upper bound for weakly-supervised method. We evaluate our method without evidence annotation and demonstrate that it can approach this upper bound.

\subsection{Self-Training MRC (\methodname)}\label{sec:our-method}
\methodname is designed to improve base MRC models via generating pseudo evidence labels for evidence extraction when golden labels are unavailable. 
\methodname works in an iterative manner, and each iteration consists of two stages. One is to learn a better base model for answer prediction and evidence labelling. The other is to obtain more precise evidence labels for the next iteration using the updated model.

At each iteration, \methodname first trains the base model with golden answers and pseudo evidence labels from the preceding iteration using the total loss as defined Equation \ref{eq:total-loss}. 
Then the trained model can predict a distribution of pseudo evidence labels for each unlabelled instance $(D,Q,A)$, and decides $\hat{E}$ as
\begin{align}
    \hat{E}&=\argmin_{E'} L_E(D,Q,E'). \label{eq:generate-noisy-label}
\end{align}
Define the confidence of a labelled instance $(D,Q,A,\hat{E})$ as
\begin{align} \nonumber
    c(D,Q,A,\hat{E})=&\exp(-L_A(D,Q,A))*\notag\\
                      &\exp(-L_E(D,Q,\hat{E}))\notag.
\end{align}
%%%%%这里的E hat也是个序列，所以是有很多序列候选，然后挑选一个出来？（那么这些候选序列如何得到的？）
$Selector$ selects the instances with the largest confidence scores whose $L_A(D,Q,A)$ and $L_E(D,Q,\hat{E})$ are smaller than the prespecified thresholds. These labelled instances will be moved from $U$ to $L$ for the next iteration.

%%%%%%%%% 加上对于算法如何启动的说明 --nyl
In the first iteration (iteration 0), the initial labeled set $L$ is set to an empty set, thus the base model is supervised only by golden answers. In this case, the evidence extractor is trained in a distant supervised manner.

The procedure of one iteration of \methodname is illustrated in Algorithm \ref{alg:algorithm}.
$\delta$ and $\epsilon$ are two thresholds (hyper-parameters). 
$sort$ operation ranks the candidate samples according to their confidence scores $s$ and returns the top-$n$ samples. $n$ varies different datasets, and \textbf{details are presented in the appendix}. 
%%%%%
% 这里加上迭代算法伪代码
% 后续还需要斟酌完善
\begin{algorithm}[htb]  
  \caption{One iteration of \methodname}  
  \label{alg:algorithm}
  \begin{algorithmic}[1]  
    \Require
      Training sets $U,L$;
      %%%%%这两个参数名字最好换一下，和前面的变量名冲突 
      %已更换--nyl
      Thresholds $\delta$ and $\epsilon$;
      Number of generated labels $n$;
      Weight of evidence loss $\eta$;
    \Ensure  
      Trained MRC model $M$; Updated training sets $U,L$;
    \State Randomly initialize $M$;
    \State Train $M$ on $U$ and $L$;
    \State Initialize $L'=\varnothing$;
    \For {each $(D,Q,A)\in U$}
        \State $l_A=L_A(D,Q,A)$;
        \State Generate $\hat{E}$ via Equation \ref{eq:generate-noisy-label};
        \State $l_{\hat{E}}=L_E(D,Q,\hat{E})$;
        \If {$l_A\leq\delta, l_{\hat{E}}\leq\epsilon$}
            \State $s=c(D,Q,A,\hat{E})$;
            \State Add $(D,Q,A,\hat{E},s)$ to $L'$;
        \EndIf
    \EndFor
    \State $L'=sort(L',n)$;
    \State $L=L\cup L', U=U\backslash L'$;
    \State \Return $M,U,L$;  
  \end{algorithmic}  
\end{algorithm}%%%%%\vspace{-2mm}

\subsection{Analysis}
%%%%%这个部分，没有实验结果，应该放在方法的后面，实验的前面？属于理论分析？ --hml
To understand why STM can improve evidence extraction and the performance of MRC, we revisit the training process and present a theoretical explanation, as inspired by \citep{selftraining-for-generation}.

%%%\textbf{Labeling Strategy.} 
% part of gains come from the use of labeling strategy, which contains some prior knowledge.
In Section \ref{sec:our-method}, we introduce the simple labeling strategy used in STM. 
If there is no sample selection, the evidence loss can be formulated as
\begin{align} \nonumber
    \mathcal{L}_{\theta^t}=-\mathbb{E}_{x\sim p(x)}\mathbb{E}_{E\sim p_{\theta^{t-1}}(E|x)}\log p_{\theta^t}(E|x),
\end{align}
where $x$ represents $(D,Q,A)$, and $\theta^t$ is the parameter of the $t$-th iteration. 
In this case, pseudo evidence labels $E$ are randomly sampled from $p_{\theta^{t-1}}(E|x)$ to guide $p_{\theta^t}(E|x)$, and therefore minimizing $\mathcal{L}_{\theta^t}$ will lead to $\theta^t=\theta^{t-1}$. 
As a matter of fact, the sample selection strategy in STM is to filter out the low-quality pseudo labels with two distribution mappings, $f$ and $g$. The optimizing target becomes 
\begin{align} \nonumber
\mathcal{L}'_{\theta^t}=-\mathbb{E}_{x\sim f(p(x))}\mathbb{E}_{E\sim g(p_{\theta^{t-1}}(E|x))}\log p_{\theta^t}(E|x).   
\end{align}
In STM, $f$ is a filter function with two pre-specified thresholds, $\delta$ and $\epsilon$. $g$ is defined as $\argmax$ (Equation \ref{eq:generate-noisy-label}). 
Compared with random sampling, our strategy tends to prevent $\theta^t$ from learning wrong knowledge from $\theta^{t-1}$. And the subsequent training might benefit from implicitly learning the strategy. In general, the strategy of STM imposes naive prior knowledge on the base models via the two distribution mappings, which may partly explain the performance gains.

% 要不要解释dropout的作用，实际上dropout和evidence label共同作用才有效果，缺少一个都可能会打折扣
% \textbf{Dropout.} At training time, the model is forced to predict the same pseudo evidence given part of input features. Such mechanism brings additional expectation over dropout noise, resulting in $\mathcal{L}''=-\mathbb{E}_{x'\sim dropout(x)}E_{E\sim p_{\theta^*}(E|x')}\log p_{\theta}(E|x')$, which trains more robust evidence extractors.

\section{Experiments}\label{sec:experiment}

\begin{table}[]
% \small
\centering
\begin{tabular}{lccc}
\toprule[1pt]
Model / Dataset         &           CoQA            & MARCO                      & BoolQ \\ \hline
\BERTMLP                &           78.0           & 70.8                      & 71.6                           \\
\BERTHAtt            &           78.8           & 71.3                      & 72.9                      \\
\BERTHAttRL                 &           79.3           & 70.3                      & 70.4                 \\
\BERTHAttRule               &           78.1           & 70.4                      & 73.8                      \\ \hline
\BERTSelfCTM    &           \textbf{80.5$^\dagger$}           & \textbf{72.3$^\ddagger$}                      & \textbf{75.2$^\dagger$}                    \\ \hline
% \BERTCoCTM      &           \textbf{80.48}           & \textbf{72.33}                      & \textbf{76.54}                 \\ \hline
\BERTHAttGold        &           82.0           &           N/A              & N/A                      \\ \bottomrule[1pt]
\end{tabular}
\caption{Classification accuracy on three Yes/No question answering datasets. N/A means there is no golden evidence label.
%%%%We applied t-test between \BERTHAtt and \BERTSelfCTM for significance test. Scores that are significantly worse than those of \BERTSelfCTM are marked with ** for $p$-value $< 0.01$.
Significance tests were conducted between \BERTSelfCTM and the best baseline of each column (t-test). $\ddagger$ means $p$-value $< 0.01$, and $\dagger$ means $p$-value $< 0.05$.
%%%%不要用experiment result这类泛泛意义上的词
}
\label{tab:result-YNQA}
%%%%%\vspace{-6mm}
\end{table}

\begin{table*}[]
\centering
% \resizebox{\textwidth}{22mm}{
% \scalebox{0.9}{
% \setlength{\tabcolsep}{1mm}{
% \small
\begin{tabular}{l|cccc|ccc|cc}
\toprule[1pt]
\multirow{3}{*}{Model / Dataset}            & \multicolumn{2}{c}{RACE-M}    & \multicolumn{2}{c|}{RACE-H}  & \multicolumn{3}{c|}{MultiRC} & \multicolumn{2}{c}{DREAM} \\
                                            & Dev         & Test            & Dev         & Test          & \multicolumn{3}{c|}{Dev}    & Dev         & Test\\
                                            & Acc         & Acc             & Acc         & Acc           & F1$_m$  & F1$_a$  & EM$_0$  & Acc         & Acc\\ \hline 
GPT+DPL                &  64.2       & 62.4            & 58.5        & 60.2               & 70.5    & 67.8    & 13.3    & 57.3        & 57.7\\
\BERTMLP                                    & 66.2       & 65.5           &  61.6      & 59.5         & 71.8   & 69.1   & 21.2   & 63.9       & 63.2       \\
\BERTHAtt                                & 67.8       & 68.2           &  62.6      & 60.4             & 70.1   & 68.1   & 19.9   & 64.2       & 62.8       \\
\BERTHAttRL                             & 68.5  & 66.9  & 62.5  & 60.0  & 72.1  & 69.5  & 21.1  & 63.1  & 63.4  \\
\BERTHAttRule                                   & 66.6       & 66.4           &  61.6      & 59.0        & 69.5   & 66.7   & 17.9   & 62.5       & 63.0       \\ 
\hline
\BERTSelfCTM                        & \textbf{69.3$^\ddagger$}       & \textbf{69.2$^\dagger$}  & \textbf{64.7$^\ddagger$}& \textbf{62.6$^\ddagger$} & \textbf{74.0$^\ddagger$}   & \textbf{70.9$^\ddagger$}   & \textbf{22.0$^\dagger$}   & \textbf{65.3$^\ddagger$}       & \textbf{65.8$^\dagger$}       \\ \hline
% \BERTCoCTM                          & \textbf{69.64}& 67.82         & 64.30       & 62.34           & \textbf{68.12}         & 66.60         & \textbf{73.99}   & \textbf{71.24}   & \textbf{24.97}   & 63.82       & 64.23       \\ \hline
\BERTHAttGold                            & N/A         & N/A             & N/A         & N/A           & 73.7   & 70.9   & 27.2   & N/A    & N/A \\
\bottomrule[1pt]
\end{tabular}
\caption{Results on three multiple-choice reading comprehension datasets. (F1$_a$: F1 score on all answer-options; F1$_m$: macro-average F1 score of all questions; EM$_0$: exact match.) 
Note that there is no golden evidence label on RACE and DREAM. 
The results for DPL (deep programming logic) are copied from \citep{MMRCselector}.
Significance tests were conducted between \BERTSelfCTM and the best baseline of each column (t-test). $\ddagger$ means $p$-value $< 0.01$, and $\dagger$ means $p$-value $< 0.05$.
}
\label{tab:result-MMRC}
%%%%%\vspace{-4mm}
\end{table*}

\subsection{Datasets}

\subsubsection{Yes/No Question Answering (YNQA)}
% 介绍各个数据集及*****数据处理细节*****
% 附加：分析一下数据集中单句evidence的比例，平均长度，平均句子数量，Yes/No数量比例，train/dev set规模

\textbf{CoQA}~\citep{CoQA} is a multi-turn conversational question answering dataset where 
questions may be incomplete and need historical context to get the answers.
%questions and their associated answers are written by two workers during the conversation on a document. 
%some questions may be incomplete and need context information from the history questions and answers. 
We extracted the Yes/No questions from CoQA, along with their histories, to form a \emph{YNQA} dataset.
% 这个说成name as YNQA合适吗？因为一共有三个YNQA数据集

\noindent\textbf{BoolQ}~\citep{BoolQ} consists of Yes/No questions from 
%real anonymized, aggregated queries issued to 
the Google search engine. Each question is accompanied by a related paragraph. 
%Most original paragraphs in BoolQ are short, making it easy to find evidence. However, in the real world, long paragraphs are more common. 
%To mimic real situations where retrieved paragraphs are long, 
We expanded each short paragraph by concatenating some randomly sampled sentences. %%%%这个会使得这个任务更复杂

\noindent\textbf{MS MARCO}~\citep{MARCO} is a large MRC dataset. 
Each question is paired with a set of reference documents and the answer may not exist in the documents.
%For each question, a set of reference documents are given to determine whether the question is answerable or not, and the corresponding answer is provided for an answerable question. Each document is annotated whether it contains enough evidence for answering a question. 
We extracted all Yes/No questions, and randomly picked some reference documents containing evidence\footnote{The evidence annotation in a document is provided by the original dataset.}. 
%to form the input documents. 
To balance the ratio of Yes and No questions, we randomly removed some questions whose answers are Yes.

\subsubsection{Multiple-choice MRC}
\noindent\textbf{RACE}~\citep{race} consists of about 28,000 passages and 100,000 questions 
%generated by human experts, which were collected 
from English exams for middle (RACE-M) and high (RACE-H) schools of China.
% We simply concatenate the passage, question and each choice as an single instance and conduct experiments on RACE-H and RACE-M independently.
The average number of sentences per passage in RACE-M and RACE-H is about 16 and 17, respectively. %%%% 这里为什么用单数形式的number？ ---nyl

\noindent\textbf{DREAM}~\citep{dream} contains 10,197 multiple-choice questions with 6,444 dialogues, collected from English examinations.
%The task requires to find the answer to a question given a dialogue session.
In DREAM, 85\% of the questions require reasoning with multiple evidential sentences. %%%%For each sample, we simply concatenated all the utterances in the current dialogue as the reference document. %%%%% 这一句话的语法有没有错误

\noindent\textbf{MultiRC}~\citep{multirc} is an MMRC dataset where the amount of correct options to each question varies from 1 to 10. 
% and we make a binary classification for each choice. 
Each question in MultiRC is annotated with evidence from its reference document.
% , which help us obtain the upper bound of \methodname. 
The average number of annotated evidence sentences for each question is 2.3. %%%%到底是多少？2，3，10，都是larger than 2！！！！be precise！！！
% the average number of sentences per passage is larger than 14.

\subsubsection{Open-domain QA (ODQA)}
\noindent\textbf{Quasar-T}~\citep{quasar} consists of 43,000 open-domain trivial questions, whose answers were extracted from ClueWeb09. For fair comparison, we retrieved 50 reference sentences from ClueWeb09 for each question the same as DSQA~\citep{DS-QA}.

\subsection{Baselines}

We compared several methods in our experiments, including some powerful base models without evidence supervision and some existing methods (*+Rule/RL/DPL/STM) which improve MRC with noisy evidence labels.
% 我们的baseline模型并不是STOA
%We selected some widely used models as baselines.
%Note that the baselines of ODQA are different from those of YNQA and MMRC, because these tasks employ different base models for \methodname. 
%%%%%%%%%% 说明会放出来源代码
\textbf{Experimental details are shown in the appendix.}
%\footnote{ The codes will be released after the review period.} \textbf{}
%due to space limit.}

\textbf{YNQA and MMRC}: (1) \textbf{\BERTMLP} utilizes a BERT encoder and an MLP answer predictor. The predictor makes classification based on the BERT representation at the position of \texttt{[CLS]}. The parameters of the BERT module were initialized from BERT-base.  (2) \textbf{\BERTHAtt} refers to the base model introduced in Section \ref{sec:bert-model}, which applies hierarchical attention over words and sentences. (3) Based on \BERTHAtt, \textbf{\BERTHAttRule} %%%%这个是不是应该叫做\BERTHAttRule更清晰？？？
supervises the evidence extractor with noisy evidence labels, which are derived from hand-crafted rules. We have explored three types of rules based on Jaccard similarity, integer linear programming (ILP)~\citep{ILPforSummary}, and inverse term frequency (ITF)~\citep{MMRCselector}, among which ITF performed best in most cases. For simplicity, we merely provided experimental results with the rule of ITF. (4) Based on \BERTHAtt, \textbf{\BERTHAttRL} trains the evidence extractor via reinforcement learning, similar to~\citep{coarse-to-fine}. And (5) another deep programming logic (DPL) method, \textbf{GPT+DPL}~\citep{MMRCselector}, is complicated and the source code is not provided, thus We directly used the results from the original paper and did not evaluate it on BERT.
%%%%这些名字全部都有BERT，
%%%%similar to 还是 the same as ... % 不一样，因为原文没有使用bert，我们为了公平对比，把bert加上去了
% For fair comparison, it is BERT rather than RNN that works as its encoder.

\textbf{ODQA}: (1) For each question, \textbf{DSQA}~\citep{DS-QA} aggregates multiple relevant paragraphs from ClueWeb09, and then infers an answer from these paragraphs. (2) \textbf{GA}~\citep{GA} and \textbf{BiDAF}~\citep{BiDAF} perform semantic matching between questions and paragraphs with attention mechanisms. And (3) \textbf{R$^3$}~\citep{R3} is a reinforcement learning method that explicitly selects the most relevant paragraph to a given question for the subsequent reading comprehension module.

\begin{table}[]
\centering
\scalebox{0.9}{
\begin{tabular}{lcc}
\toprule[1pt]
Model           & EM            & F1   \\ \hline
GA~\citep{GA}           & 26.4          & 26.4\\
BiDAF~\citep{BiDAF}    & 25.9          & 28.5\\
R$^3$~\citep{R3}        & 35.3          & 41.7\\ \hline
DSQA~\citep{DS-QA}      & 40.7          & 47.6\\ 
\quad +distant supervision       & 41.7          & 48.7\\
\quad +\methodname       & \textbf{41.8$^\dagger$}          & \textbf{49.2$^\dagger$}\\
\bottomrule[1pt]
\end{tabular}}
\caption{Experimental results on the test set of Quasar-T. R$^3$ is a RL-based method.  Results of GA, BiDAF and R$^3$ are copied from \citep{DS-QA}.
%%%We applied t-test between DSQA and DSQA+\methodname for significance test. Scores that are significantly worse than those of DSQA+\methodname are marked with * for $p$-value $< 0.05$.
DSQA+STM outperforms the best baseline (DSQA+DS) significantly (t-test, $p$-value $< 0.05$, DS=distant supervision).}
\label{tab:result-Quasar}
%%%%%\vspace{-4mm}
\end{table}

\subsection{Main Results}

\subsubsection{Yes/No Question Answering}

Table \ref{tab:result-YNQA} shows the results on the three YNQA datasets. We merely reported the classification accuracy on the development sets since the test sets are unavailable. %We adopted classification accuracy as the metric.

%%%YNQA is a classification task in nature, so that we chose classification accuracy as the metric. %%%%废话太多
\BERTSelfCTM outperformed all the baselines, which demonstrates the effectiveness of our method. %%%%%The larger margin of \methodname on BoolQ shows that evidence finding is relatively simple for BoolQ.
Compared with \BERTMLP, \BERTHAtt achieved better performance on all the three datasets, indicating that distant supervision on evidence extraction can benefit Yes-No question answering. 
However, %%%%%什么叫做没有效果？？相对什么没有提高？？？还是什么？？？
compared with \BERTHAtt, \BERTHAttRL made no improvement on MARCO and BoolQ, possibly due to the high variance in training. 
Similarly, \BERTHAttRule performed worse than \BERTHAtt on CoQA and MARCO, implying that it is more difficult for the rule-based methods (inverse term frequency) %%% what is this? who can remember?
to find correct evidence in these two datasets. 
%%%%%%%%%%%% 要不要在这里说一下，STM得益于好的稳定性和迁移性 ---nyl
In contrast, our method \BERTSelfCTM is more general and performed the best on all datasets. \BERTSelfCTM achieved comparable performance with \BERTHAttGold which stands for the upper bound by providing golden evidence labels, indicating that the effectiveness of noisy labels in our method. 
%There is no upper bound performance for \BERTSelfCTM on BoolQ and MARCO because golden labels are not available in the two datasets. 
%%%%%%以上把顺序调整一下；
%%%%先说最重要的主要的结论，（自己的方法）
%%%%然后才是细枝末节的

\subsubsection{Multiple-choice MRC}

Table \ref{tab:result-MMRC} shows the experimental results on the three MMRC datasets. We adopt the metrics from the referred papers. \methodname improved \BERTHAtt consistently %%%我不觉得提高很大
on RACE-H, MultiRC and DREAM in terms of all the metrics. %%%%什么地方有很大提高？不要使劲强调接近gold
% Similarly, RACE-H and DREAM both benefit a lot from \methodname, with gain up to $1.9$. %%%这个句子主语是谁呢？为什么是数据集benefit from？？？
However, the improvement on RACE-M is limited ($1.0$ gain on the test sets). The reason may be that RACE-M is much simpler than RACE-H, and thus, it is not challenging for the evidence extractor of \BERTHAtt to find the correct evidence on RACE-M. %%%%大家都是distant，谁跟谁的比较；搞清楚咱们在讨论的对象是什么？？？
%%%这个最后一句应该加一条in RACE-M之类的限制，不然有指代不明的问题

\subsubsection{Open-domain Question Answering}

%%%%这段重新写一下吧
Table \ref{tab:result-Quasar} shows the exact match scores and F1 scores on Quasar-T. 
%For the metrics, we adopt the exact match score and F1 score, computed by comparing a prediction and a golden answer. 
Distant evidence supervision (DS) indicates whether a passage contains the answer text.
Compared with the base models DSQA and DSQA+DS, DSQA+\methodname achieved better performance in both metrics, which
verifies that DSQA can also benefit from Self-Training.
Our method is general and can improve both lightweight and heavyweight models, like LSTM-based and BERT-based models, in different tasks. 
%verifies that our method is generalizable to non-BERT models. %%%%%%% 这个non用法对吗
%In other words, the effectiveness of \methodname is not dependent on pre-trained language models and the structure of Transformer.
%%%%这个比较没有意思，重点相比原有的方法提高多少？
%%%%There is still room for further improvement if we apply more effective Co-Training methods.

\begin{table}[]
\centering
\setlength{\tabcolsep}{0.5mm}{
% \scalebox{0.8}{
\small
\begin{tabular}{l|c|cccccc}
\toprule[1pt] %%%% coqa每个样本只有一个evidence sentence，所以只计算了accuracy
\multirow{2}{*}{Model/Dataset}& CoQA     & \multicolumn{6}{c}{MultiRC} \\
                 & P@1      & R@1    & R@2    & R@3    & P@1    & P@2    & P@3     \\ \hline             %& p@4     & p@5     \\ \hline
\BERTHAtt        & 20.0    & 28.2  & 49.8  & 62.5  & 62.3  & 55.2  & 46.6   \\              %& 40.22   & 35.24   \\ \hline
% \BERTCoCTM+View0 & 25.08    & 31.48  & 55.16  & 69.17  & 69.57  & 61.28  & 51.76   \\                    %& 43.57   & 37.80   \\
% \BERTCoCTM+View1 & 27.30    & 31.00  & 55.47  & 68.53  & 68.63  & 61.49  & 51.28   \\                    %& 43.31   & 37.57   \\
\quad +RL       & 5.2             &  10.5          & 22.3           & 32.9           & 24.0           & 25.3           & 24.7   \\ 
\quad +Rule     & 38.4             & 32.4           & 53.6           & 65.1           & 71.8           & 59.6           & 48.7   \\ 
\quad +\methodname(iter 1)     & 32.7             & 32.8           & 57.1           & 70.1           & 72.2           & 63.3           & 52.5   \\
\quad +\methodname(iter 2)     & 37.3             & \textbf{32.9}  & \textbf{58.0}  & \textbf{71.3}  & \textbf{72.7}  & \textbf{64.4}  & \textbf{53.5}   \\
\quad +\methodname(iter 3)     & \textbf{39.9}    & 31.4           & 55.3           & 68.8           & 69.5           & 61.6           & 51.6   \\
\hline
\BERTHAttGold    & 53.6    & 33.7  & 59.5  & 73.4  & 74.5  & 65.9  & 54.8   \\ \bottomrule[1pt]   %& 46.25   & 39.54   \\   \bottomrule[2pt]
\end{tabular}}
\caption{Evidence extraction evaluation on the development sets of CoQA and MultiRC. P@$k$ / R@$k$ represent precision / recall of the generated evidence labels, respectively for top $k$ predicted evidence sentences.}
\label{tab:extractor-performance}
%%%%%\vspace{-5mm}
\end{table}

\subsection{Performance of Evidence Extraction}

%We claimed that \methodname can improve evidence extraction to help a base model achieve better performance on MRC. 
To evaluate the performance of \methodname on evidence extraction, we validated the evidence labels generated by several methods on the development sets of CoQA and MultiRC. 
% 待补充：解释p@k/r@k/Acc
Considering that the evidence of each question in MultiRC is a set of sentences, we adopted $precision@k$ and $recall@k$ as the metrics for MultiRC, which represent the precision and recall of the generated evidence labels, respectively, when $k$ sentences are predicted as evidence. 
We adopted only $precision@1$ as the metric for CoQA as this dataset provides each question with one golden evidence sentence. %%%%这个有标注 
%%%%%既然有golden为什么不直接算top 1；top 2； top 3 precision；或者阈值截断算acc？？？
%%% 没办法计算top 1；top 2； top 3 precision，因为每个multirc样本具有多个evidence sentences，只能去算F1；但是HAtt这个baseline也没有被设计成可以生成多个evidence sentences，这样可能就不适合F1这种评判方式了。而且从直观上讲，我们既然claim说是soft evidence extraction，那么使用这种soft的评测方式应该更直观合理吧。---nyl

Table \ref{tab:extractor-performance} shows the performance of five methods for evidence labeling on the CoQA and MultiRC development sets. It can be seen that \BERTSelfCTM outperformed the base model \BERTHAtt by a large margin in terms of all the metrics. 
As a result, the evidence extractor augmented with \methodname provided more evidential information for the answer predictor, which may explain the improvements of \BERTSelfCTM on the two datasets. 

\subsection{Analysis on Error Propagation}

\begin{figure}[ht]
    \centering
    \includegraphics[width=0.5\textwidth]{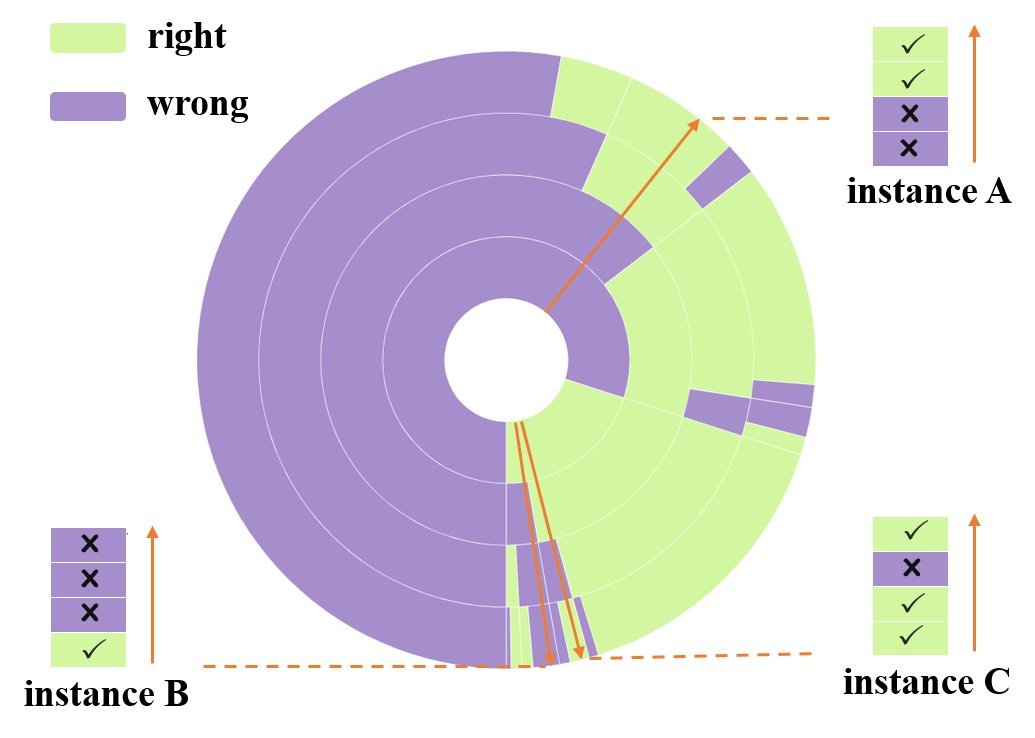}
    % \vspace{-3mm}
    \caption{Evolution of evidence predictions on the development set of CoQA. From the inside to the outside, the four rings correspond to \BERTHAtt (iteration 0) and \BERTSelfCTM (iteration 1, 2, 3), respectively. %%%%%If an instance is predicted correctly, the corresponding radius is marked as green, otherwise as purple. 
    }
    \label{fig:case-statistics}
    %%%%%\vspace{-4mm}
\end{figure}

To examine whether error propagation exists and how severe it is in \methodname, we visualized the evolution of evidence predictions on the development set of CoQA (Figure \ref{fig:case-statistics}). 
From the inside to the outside, the four rings 
show the statistic results of the evidence predicted by \BERTHAtt (iteration 0) and \BERTSelfCTM (iteration 1, 2, 3).
%correspond to \BERTHAtt (iteration 0) and \BERTSelfCTM (iteration 1, 2, 3), respectively. 
Each ring is composed of all the instances from the development set of CoQA, and each radius corresponds to one sample. If the evidence of an instance is predicted correctly, %%%%是evidence对，还是答案对???
the corresponding radius is marked in green, otherwise in purple. 
%The numbers of instances in five regions are presented in the figure.
\textbf{Two examples are shown in the appendix due to space limit.}

\textbf{Self-correction.} 
As the innermost ring shows, about $80\%$ of the evidence predicted by \BERTHAtt (iter 0) was incorrect. However, the proportion of wrong instances reduced to $60\%$ after self-training (iter 3).
More concretely, $27\%$ of the wrong predictions were gradually corrected with high confidence within three self-training iterations, as exemplified by {\it instance A} in Figure \ref{fig:case-statistics}.
%In iteration 0, there were 1295 instances (region D) predicted with incorrect evidence, which occupied about $80\%$ of the test samples. %%%predict answer or evidence???
%It is clear that the base model was gradually correcting itself with \methodname iterations. As a result, $27\%$ of the wrong predictions from region D were corrected (region E) with high confidence until iteration 3.%%%27的分母是什么

\textbf{Error propagation.} 
We observed that $4\%$ of the evidence was mistakenly revised by \methodname, as exemplified by {\it instance B} in Figure \ref{fig:case-statistics}. In such a case, the incorrect predictions are likely to be retained in the next iteration. But almost $50\%$ of such mistakes were finally corrected during the subsequent iterations like {\it instance C}. This observation shows that \methodname can prevent error propagation to avoid catastrophic failure.
%Region B represents a particular type of evidence predictions, which were mistakenly revised by \methodname. In this case, the incorrect predictions are likely to be retained in the next iteration, which is called error propagation. Another observation is that a nontrivial proportion ($50\%$) of such mistakes were corrected in the subsequent iterations (region D). This observation shows that the self-correction of \methodname prevented error propagation from being catastrophic.

%%%%这个展示方式还是不清楚，不如画个表???
%%%%这些数值不解释的话，不清楚什么意思

\subsection{Improvement Over Stronger Pretrained Models}

To evaluate the improvement of \methodname over stronger pretrained models, we employed RoBERTa-large~\citep{roberta} as the encoder in the base model. Table \ref{tab:roberta-performance} shows the results on CoQA. \methodname significantly improved the evidence extraction (Evi. Acc) of the base model. However, the improvement on answer prediction (Ans. Acc) is marginal. One reason is that RoBERTa-HA achieved so high performance that there was limited room to improve. Another possible explanation is that evidence information is not important for such stronger models to generate answers. In other words, they may be more adept at exploiting data bias to make answer prediction. In comparison, weaker pretrained models, such as BERT-base, can benefit from evidence information due to their weaker ability to exploit data bias.

\begin{table}[]
\centering
\begin{tabular}{lcc}
\toprule[1pt]
Model/Metric                                                         & Ans. Acc & Evi. Acc \\ \hline
RoBERTa-HA                                               & 92.6     &     13.8     \\
RoBERTa-HA+\methodname & \textbf{92.7}     &       \textbf{19.3}(+40\%)   \\ 
% RoBERTa-HA+Gold                                          & 92.6     &     \textbf{59.2}     \\ 
\bottomrule[1pt]
\end{tabular}
\caption{Answer prediction accuracy (Ans. Acc) and evidence extraction accuracy (Evi. Acc) on the development set of CoQA.}
\label{tab:roberta-performance}
\end{table}

\iffalse
\subsection{Correlation Analysis}

\begin{table}[]
\centering
\begin{tabular}{lc}
\toprule[1pt]
Model          & Correlation \\ \hline
BERT+HAtt+RL   & 0.013       \\
BERT+HAtt+Rule & 0.139       \\
BERT+HAtt+Gold & 0.149       \\ \hline
BERT+HAtt+STM  &             \\
\qquad\quad iteration 0    & 0.146       \\
\qquad\quad iteration 1    & 0.163       \\
\qquad\quad iteration 2    & 0.172       \\
\qquad\quad iteration 3    & 0.204       \\ \bottomrule[1pt]
\end{tabular}
\end{table}

\subsection{Further Improvement}
\fi

\section{Conclusion and Future Work}

We present an iterative self-training method (STM) to improve MRC models with soft evidence extraction, when golden evidence labels are unavailable. 
In this iterative method, we train the base model with golden answers and pseudo evidence labels. The updated model then generates new pseudo evidence labels, which can be used as additional supervision in the next iteration. 
%This iterative process terminates until no further improvement can be obtained. 
%Our method is much effective and general.
Experiments results show that our proposed method consistently improves the base models in seven datasets for three MRC tasks, and that better evidence extraction indeed enhances the final performance of MRC. 

As future work, we plan to extend our method to other NLP tasks which rely on evidence finding, such as natural language inference. %%%%We will also investigate the interpretability of \methodname via the generated evidence labels.
%%%%%I do not like them at all ---hml
%%\begin{enumerate}
%%    \item In this work, we merely employ the simplest Co-Training method. A more effective Co-Training method can further improve the effectiveness of \methodname. We will design a specialized Co-Training algorithm for \methodname to enhance the performance.
%%    \item Rule based labeling methods and reinforcement learning methods are not mutual exclusion to \methodname. We will explore how to incorporate these three types of methods to get further improvement.
%%%\end{enumerate}

\section*{Acknowledgments}
This work was jointly supported by the NSFC projects (Key project with No. 61936010 and regular project with No. 61876096), and
the National Key R\&D Program of China (Grant No. 2018YFC0830200). We thank THUNUS NExT Joint-Lab for the support.

\bibliographystyle{acl_natbib}
% \bibliography{acl2020}

\appendix

\section{Case Study}

In Section 4.5 of the main paper, we provide a quantitative analysis of the evolution of evidence predictions, and draw two conclusions: (1) STM can help the base model to correct itself; (2) Error propagation will not result in catastrophic failure, though exists. 

\begin{figure}[ht]
    \centering
    \includegraphics[width=0.5\textwidth]{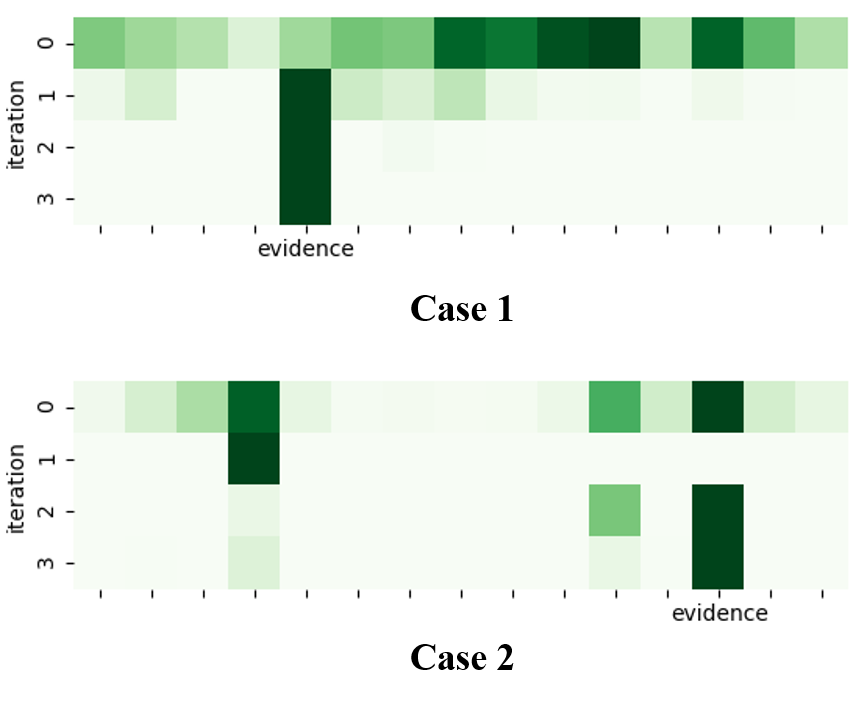}
    \caption{Weight distribution of the two cases from the sentence-level attention. }
    \label{fig:appendix-case}
\end{figure}

To help understand these two conclusions, we provide two corresponding cases from the development set of CoQA~\citep{CoQA}. The original instances are shown in Table \ref{tab:appendix-example}, and the weight distribution from the sentence-level attention is shown in Figure \ref{fig:appendix-case}. 
In case 1, BERT-HA made wrong evidence prediction, while STM revised it subsequently, which shows the ability of self-correction. 
In case 2, BERT-HA first selected the correct evidence with high confidence. However, in the iteration 1, BERT-HA with STM was distracted by another plausible sentence. Instead of insisting on the incorrect prediction, STM led BERT-HA back to the right way, which shows that error propagation is not catastrophic.

\section{Hyper-Parameters for Self-Training}

We implemented BERT-HA with BERT-base from a commonly used library\footnote{https://github.com/huggingface/transformers}, and directly used the original source code of DSQA\footnote{https://github.com/thunlp/OpenQA}~\citep{DS-QA}. All the codes and datasets will be released after the review period. 
The hyper-parameters used in BERT-HA and BERT-HA+STM are shown in Table \ref{tab:hyper-parameters}.

\begin{table*}[]
\centering
\begin{tabular}{|p{\textwidth}|}
\hline
\textbf{(Case 1)}\\
\textbf{Passage:}\\
...(3)"Why don't you tackle Indian River, Daylight?" (4)Harper advised, at parting. \textcolor{red}{(5)"There's whole slathers of creeks and draws draining in up there, and somewhere gold just crying to be found.} (6)That's my hunch. (7)There's a big strike coming, and Indian River ain't going to be a million miles away. 
(8)"And the place is swarming with moose," Joe Ladue added.
\textcolor{blue}{(9)"Bob Henderson's up there somewhere, been there three years now, swearing something big is going to happen, living off'n straight moose and prospecting around like a crazy man."
(10)Daylight decided to go Indian River a flutter, as he expressed it; but Elijah could not be persuaded into accompanying him.
Elijah's soul had been seared by famine, and he was obsessed by fear of repeating the experience.
(11)"I jest can't bear to separate from grub," he explained.}
(12)"I know it's downright foolishness, but I jest can't help it..."\\
\textbf{Question:} Are there many bodies of water there?\\
\textbf{Answer:} No\\
\hline
\textbf{(Case 2)}\\
\textbf{Passage:}\\
(1)If you live in the United States, you can't have a full-time job until you are 16 years old.
(2)At 14 or 15, you work part-time after school or on weekends, and during summer vacation you can work 40 hours each week.
(3)Does all that mean that if you are younger than 14, you can't make your own money?
(4)Of course not!
\textcolor{blue}{(5)Kids from 10-13 years of age can make money by doing lots of things.}
(6)Valerie, 11, told us that she made money by cleaning up other people's yards.
...(11)Kids can learn lots of things from making money. (12)By working to make your own money, you are learning the skills you will need in life. \textcolor{red}{(13)These skills can include things like how to get along with others, how to use technology and how to use your time wisely.} (14)Some people think that asking for money is a lot easier than making it; however, if you can make your own money, you don't have to depend on anyone else...\\
\textbf{Question:} Can they learn time management?\\
\textbf{Answer:} No\\
\hline
\end{tabular}
\caption{Examples from the development set of CoQA. Evidential sentences in red in reference passages are crucial to answer the questions. Sentences in blue are distracting as Figure \ref{fig:appendix-case} shows.}
\label{tab:appendix-example}
\end{table*}

\begin{table*}[]
\centering
% \scalebox{0.9}{
% \setlength{\tabcolsep}{1mm}{
\begin{tabular}{lccccccc}
\toprule[1pt]
Dataset             & RACE-H    & RACE-M    & DREAM     & MultiRC   &  CoQA     & MARCO     & BoolQ   \\ \hline
$L_{\max}$      & 380       & 380       & 512       & 512       & 512       & 480       & 512 \\
learning rate       & 5e-5$^\clubsuit$/4e-5$^\spadesuit$ & 5e-5$^\clubsuit$/4e-5$^\spadesuit$ & 2e-5 & 2e-5 & 2e-5 & 2e-5 & 3e-5\\
epoch               & 3         & 3         & 5         & 8         & 3         & 2$^\clubsuit$/3$^\spadesuit$       & 4\\
$\eta$  & 0.8       & 0.8       & 0.8       & 0.8       & 0.8       & 0.8       & 0.8 \\
batch size          & 32        & 32        & 32        & 32        & 6         & 8         & 6\\ 
$\epsilon$    & 0.5       & 0.5       & 0.5       & 0.5       & 0.6       & 0.5       & 0.5 \\ 
$\delta$     & 0.9       & 0.9       & 0.8       & 0.8       & 0.9       & 0.9       & 0.7 \\ 
$n$               & 40000     & 10000     & 3000      & 2000      & 1500      & 1000      & 500 \\
$K_{\max}$    & 2         & 3         & 4         & 3         & 1         & 1         & 1 \\ \bottomrule[1pt]
\end{tabular}
\caption{Hyper-parameters marked with $\clubsuit$/$\spadesuit$ are used in BERT-HA/BERT-HA+STM, respectively. Other unmarked hyper-parameters are shared by these two models.}
\label{tab:hyper-parameters}
\end{table*}

\end{document}